\ifx\pdfminorversion\undefined\else\pdfminorversion=6\fi
\documentclass[11pt]{article}

\usepackage[a4paper,textwidth=170mm,textheight=250mm,centering]{geometry}
\usepackage[T1]{fontenc}
\usepackage[utf8]{inputenc}
\usepackage{lmodern}
\usepackage{graphicx}
\usepackage{float}
\usepackage{booktabs}
\usepackage{array}
\usepackage{xcolor}
\usepackage{amsmath}
\usepackage[compress,nospace]{cite}
\usepackage[hidelinks]{hyperref}
\usepackage{url}

\newcommand{\dataset}{PhononBench-MP40}
\newcommand{\Stable}{Stable}
\newcommand{\Unstable}{unstable}
\newcommand{\wmin}{\omega_{\min}}

\begin{document}
\title{\dataset: a spectrum-resolved benchmark dataset for phonon stability}

\author{Wen-Kao Li$^{1,\dagger}$, Ze-Feng Gao$^{1,\dagger,*}$,  Zhong-Yi Lu$^{1,*}$\\[2pt]
\small $^1$School of Physics and Key Laboratory of Quantum State Construction \\
\small and Manipulation (Ministry of Education), Renmin University of China, Beijing 100872, China\\[2pt]
\small ${}^{\dagger}$These authors contributed equally.\\
\small ${}^{*}$e-mail: zfgao@ruc.edu.cn; zlu@ruc.edu.cn}
\date{\today}
\maketitle

\begin{abstract}
Imaginary phonon modes remain a practical bottleneck in computational materials screening because otherwise plausible structures can be locally dynamically unstable under a chosen workflow. Here we present \dataset, a spectrum-resolved benchmark dataset of Materials Project-derived crystals for workflow-defined phonon stability. The dataset starts from 47,969 MP40 workflow tasks and provides 46,899 completed records with paired stability labels and local phonopy YAML spectra, including 16,683 \Stable\ records and 30,216 completed-phonon \Unstable\ records. A further 1,067 relaxation failures are reported separately rather than merged into the completed phonon denominator. The release centers on the local YAML spectrum: the stability label, the lowest sampled frequency and any threshold-dependent relabeling are derived from that spectrum. The dataset is openly available through Science Data Bank at \url{https://doi.org/10.57760/sciencedb.38735}. A companion GitHub repository provides the calculation code and lightweight access utilities. \dataset\ provides an auditable reference for workflow-defined stability classification, minimum-frequency analysis, threshold studies and failure-aware triage, while keeping the reference workflow, data schema and interpretation boundaries explicit.
\end{abstract}

\noindent\textbf{Keywords:} phonon stability; benchmark dataset; phonon spectra; materials informatics

\section{Introduction}

Large-scale crystal generation and screening have changed the rate at which candidate materials are proposed \cite{Xie2022CDVAE,Jiao2023DiffCSP,Merchant2023GNoME,Zeni2025MatterGen}. The bottleneck is no longer only how to generate structures, but how to decide which of them deserve expensive follow-up. Formation energy, charge balance, symmetry and structural validity are useful filters, but they do not by themselves establish local dynamical stability. Within a specified interatomic-potential and phonon workflow, the relevant question is sharper: does the relaxed structure have a sampled phonon mode with negative curvature, i.e. an imaginary frequency? This makes phonon stability a natural target for reusable computed data and for model evaluation.

The field already has extensive computed-data infrastructure. Materials Project, AFLOW, OQMD, JARVIS, NOMAD and Materials Cloud organize large computational materials spaces \cite{Jain2013MaterialsProject,Curtarolo2012AFLOW,Saal2013OQMD,Kirklin2015OQMD,Choudhary2020JARVIS,Draxl2018NOMAD,Talirz2020MaterialsCloud}. Workflow and analysis tools such as pymatgen, ASE, FireWorks, atomate and AiiDA support reproducible workflows \cite{Ong2013Pymatgen,Larsen2017ASE,Jain2015FireWorks,Mathew2017Atomate,Pizzi2016AiiDA}. High-throughput phonon resources and software have established practical routes to vibrational and thermal properties, including DFPT phonons, AFLOW-AAPL, ShengBTE, ALAMODE, TDEP, SCAILD and recent anharmonic-phonon databases \cite{Petretto2018DFPTPhonons,Plata2017AFLOWAAPL,Carrete2014HalfHeusler,Mounet2018TwoDMaterials,Li2014ShengBTE,Tadano2014ALAMODE,Hellman2011TDEP,Souvatzis2009SCAILD,Ohnishi2026Phonix}. Most existing open phonon datasets contain spectra or vibrational properties for thousands to tens of thousands of materials. \dataset\ extends this scale to 46,899 completed MP-derived phonon records, while preserving the spectral object behind each stability label.

This design matters because a binary stability flag is convenient but incomplete. A shallow soft mode, a deep imaginary branch and a failed calculation are different records for both physics and workflow auditing. Collapsing them into one undifferentiated ``unstable'' class hides the evidence behind the decision and obscures the denominator used in later statistics. \dataset\ keeps these cases separate. It pairs labels with local phonopy YAML spectra, fixes the completed label+YAML cohort, and reports missing-spectrum relaxation failures outside the completed phonon denominator. The result is both a data release and a benchmark target: users can work with a fixed PhononBench-defined label, but can also inspect the spectrum, extract a scalar frequency diagnostic, or move the stability threshold for their own application.

\section{Dataset construction and workflow}

\begin{figure}[t]
  \centering
  \includegraphics[width=0.98\textwidth]{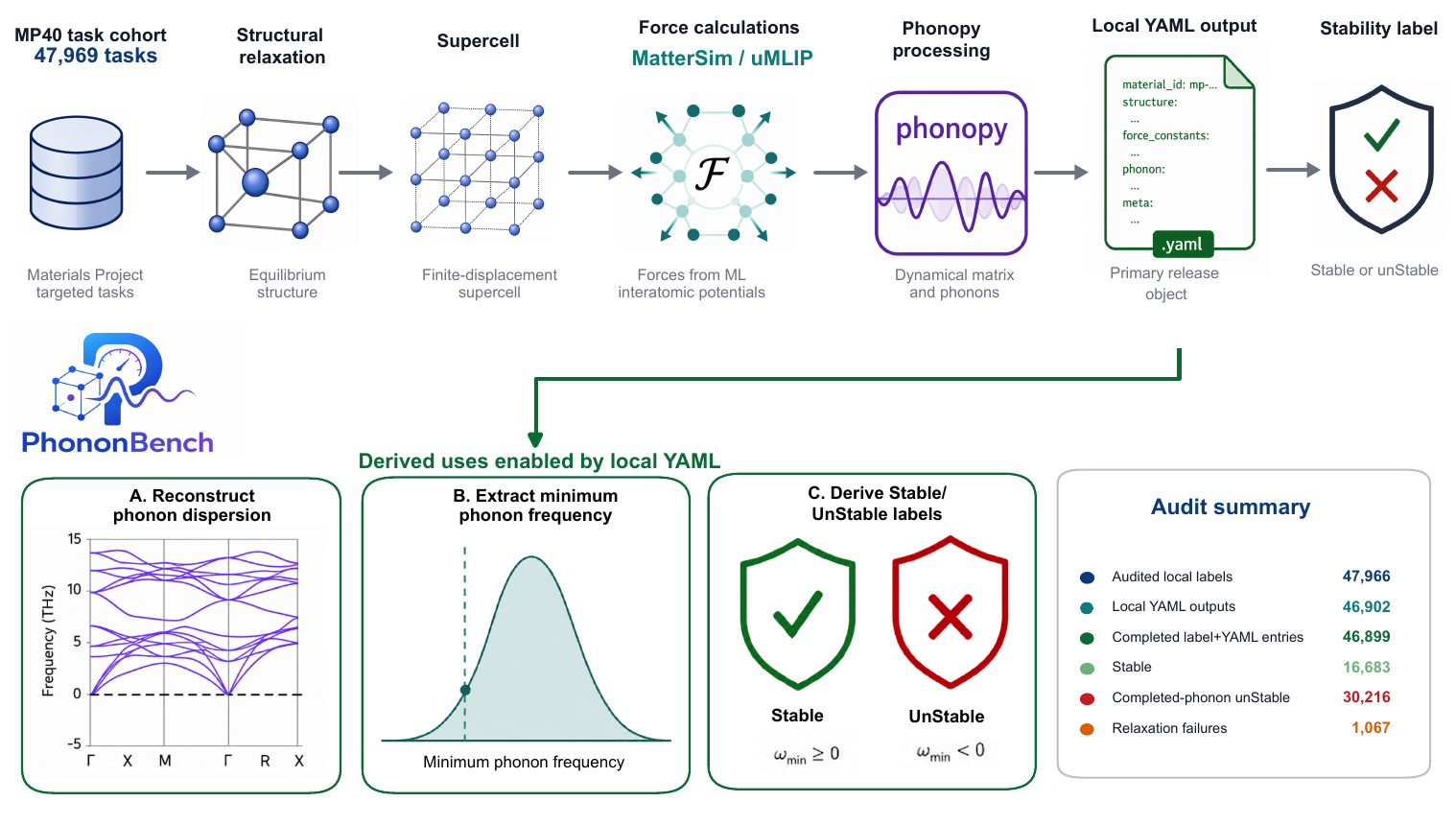}
  \caption{\label{fig:workflow}\textbf{Workflow and release object of \dataset.} The dataset starts from 47,969 MP40 tasks and follows the PhononBench route through structural relaxation, finite-displacement supercell construction, MatterSim/uMLIP force evaluation and phonopy processing. The local YAML output is the primary release object because it enables dispersion reconstruction, minimum-frequency extraction and stability-label derivation. The audit summary reports 47,966 audited local labels, 46,902 local YAML outputs, 46,899 completed label+YAML entries, 16,683 \Stable\ entries, 30,216 completed-phonon \Unstable\ entries and 1,067 relaxation failures.}
\end{figure}

Figure~\ref{fig:workflow} summarizes the construction. The starting point is an audited MP40 task cohort containing 47,969 recorded tasks within the local $N_{\mathrm{atoms}}\le 40$ workflow scope. Atom-count windows drawn from Materials Project are common in crystal-generation benchmarks, including MP-20, MP-21--40 and MP-40 settings \cite{Xie2022CDVAE,Xiao2023SLICES,Park2025ChemEleon}. Here the cutoff also has a direct phonon-workflow meaning. Under the $2\times2\times2$ finite-displacement supercell used in the workflow, a 40-atom recorded cell corresponds to an up-to-320-atom force-evaluation cell before symmetry reduction. The MP40 cutoff therefore defines a practical release domain: it includes many inorganic prototypes, ordered derivatives and multicomponent cells, while keeping the large-scale finite-displacement calculation computationally controlled. It is a benchmark-domain definition rather than a physical stability boundary.

Each task follows the PhononBench route. Structures are relaxed, finite-displacement supercells are constructed, forces are evaluated with MatterSim/uMLIP, and phonopy is used to build the dynamical matrix and phonon outputs \cite{Togo2015Phonopy,Togo2023Phonopy,Yang2024MatterSim,Han2025PhononBench}. High-symmetry paths are generated through the seekpath/spglib ecosystem \cite{Hinuma2017Seekpath,Togo2024Spglib}. The top row of Fig.~\ref{fig:workflow} is therefore shown as an explicit workflow rather than as a stand-alone label source: the reference target is a computed phonon result under a specified route.

The primary release object is the local phonopy YAML file. It contains the spectral information needed to reconstruct the dispersion, extract the lowest sampled phonon frequency and derive a Stable/unstable decision. The label is thus a derived view of the spectrum. A completed record in \dataset\ is defined by the intersection of two local objects: a recovered workflow stability label and a matching local YAML output. This rule yields 46,899 completed label+YAML records from 47,966 audited local labels and 46,902 local YAML outputs. Within the completed cohort, 16,683 records are \Stable\ and 30,216 records are completed-phonon \Unstable. The audit also reports 1,067 relaxation failures separately, because no matching local YAML spectrum is available for those records. This separation is essential for denominator-aware reuse: a completed imaginary-mode spectrum and a missing-spectrum relaxation failure answer different scientific and computational questions. Release-file definitions, common fields, workflow settings and the label rule are provided in the Supplementary Information, Secs.~S1--S3.

\begin{table}[t]
  \centering
  \caption{\label{tab:release}\textbf{Cohort accounting.} Counts for the MP40 audit and completed spectral cohort.}
  \small
  \begin{tabular}{@{}lr@{}}
    \toprule
    Quantity & Count \\
    \midrule
    Audited MP40 tasks & 47,969 \\
    Recovered workflow labels & 47,966 \\
    Recovered phonopy YAML outputs & 46,902 \\
    Completed label+YAML records & 46,899 \\
    \Stable\ completed records & 16,683 \\
    Completed-phonon \Unstable\ records & 30,216 \\
    Relaxation failures without YAML & 1,067 \\
    \bottomrule
  \end{tabular}
\end{table}

\section{Data records}

The released data are organized to make the Stable/unstable decision traceable. The archival Science Data Bank record contains the completed label table, the failure-accounting table, the reference split table, release metadata and the phonopy YAML spectra for completed records. Table~\ref{tab:data-records} summarizes the data objects that define the public release. The completed table is the main machine-learning table: it links each completed record to a structure identifier, formula, space group, unit-cell atom count, workflow label, completed label, YAML path, $\wmin$ value and split assignment. The failure table is kept separate because a relaxation failure without a YAML spectrum is a workflow outcome, not a completed phonon-instability label.

This separation is important for reuse. Users who need a fixed binary target can train on the completed label+YAML cohort. Users who need more physically resolved targets can reconstruct the band frequencies from the YAML files, calculate $\wmin$, or impose an application-specific threshold. Users interested in high-throughput workflow robustness can include the 1,067 relaxation failures in a separate failure-aware triage task. The file schema, field definitions, split construction and example parsing workflow are given in the Supplementary Information, Secs.~S1, S2, S5 and S9.

\begin{table}[t]
  \centering
  \caption{\label{tab:data-records}\textbf{Released data records.} The public release is defined by tabular metadata plus phonopy YAML spectra. Counts refer to the release described in this work.}
  \small
  \begin{tabular}{@{}p{0.24\textwidth}rp{0.50\textwidth}@{}}
    \toprule
    Data object & Records & Main use \\
    \midrule
    Audited MP40 task table & 47,969 & Defines the recorded MP-derived task scope and denominator for audit accounting. \\
    Completed label+YAML table & 46,899 & Main spectral cohort linking structure identifiers, workflow labels, completed labels, YAML paths, $\wmin$ values and split assignments. \\
    Phonopy YAML spectra & 46,899 & Primary scientific object for reconstructing dispersions, extracting frequency diagnostics and re-deriving threshold-dependent labels. \\
    Relaxation-failure table & 1,067 & Separate workflow-failure cohort for failure-aware triage and denominator-aware reporting. \\
    Reference split table & 46,899 & Formula-group train/validation/test assignment for benchmark comparisons without exact-formula leakage. \\
    Metadata and integrity files & -- & Version, repository, file-integrity and reuse information for reproducible access. \\
    \bottomrule
  \end{tabular}
\end{table}

\section{Spectra behind the labels}

The completed stability label is derived from high-symmetry-path band frequencies. A completed entry is labeled \Unstable\ if any sampled band frequency is below $-10^{-3}$ THz; otherwise it is labeled \Stable. This threshold turns a phonon dispersion into a machine-readable target, but the spectrum is the more informative object. We use $\wmin$ as the lowest sampled frequency along the released path. For negative values, $\wmin$ measures the severity of an imaginary branch under the reference workflow; near-zero values identify threshold-sensitive cases. It is a compact diagnostic of the YAML spectrum, not a replacement for the full dispersion and not a model-independent stability law.

\begin{figure}[H]
  \centering
  \includegraphics[width=0.98\textwidth]{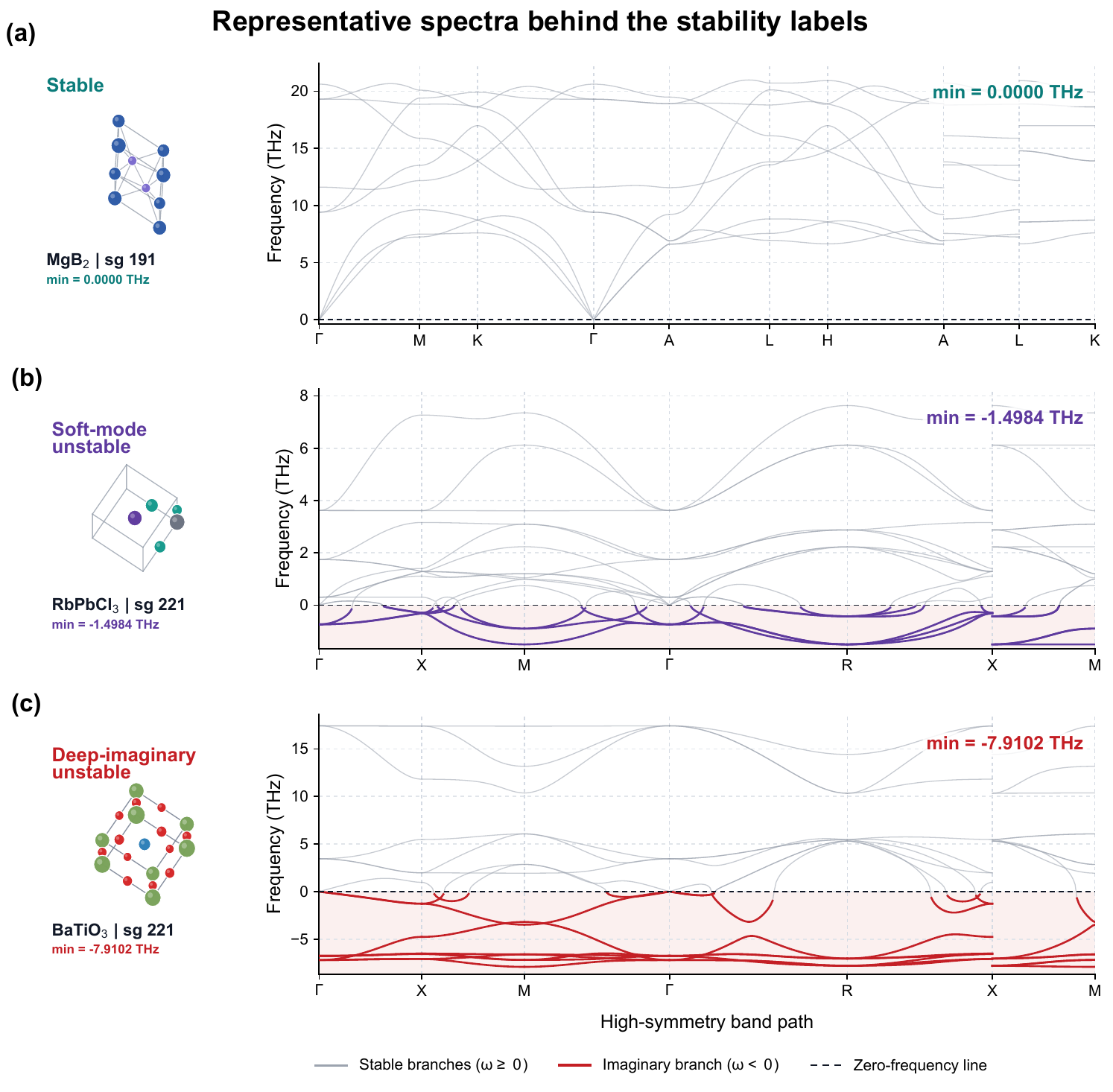}
  \caption{\label{fig:spectra}\textbf{Representative phonon spectra behind the stability labels.} Three YAML-derived dispersions illustrate the label rule and the information retained beyond the binary decision. \textbf{a,} MgB$_2$ is labeled \Stable, with $\wmin=0.0000$ THz within the adopted tolerance. \textbf{b,} RbPbCl$_3$ is a soft-mode completed-phonon \Unstable\ example with $\wmin=-1.4984$ THz. \textbf{c,} BaTiO$_3$ is a deep-imaginary completed-phonon \Unstable\ example with $\wmin=-7.9102$ THz. Grey branches denote non-negative modes, colored branches denote imaginary modes and the dashed line marks zero frequency.}
\end{figure}

Figure~\ref{fig:spectra} illustrates why the spectral object matters. MgB$_2$ is labeled \Stable\ within the adopted tolerance, with $\wmin=0.0000$ THz. RbPbCl$_3$ is a completed-phonon \Unstable\ example with a soft mode and $\wmin=-1.4984$ THz. Such shallow imaginary modes are often the cases most sensitive to relaxation details, threshold choice, supercell construction and low-symmetry distortions. BaTiO$_3$ is also \Unstable, but with a much deeper imaginary branch, $\wmin=-7.9102$ THz, indicating a stronger negative-curvature signal under the same reference route. These records share the same binary decision, yet the underlying spectra carry different physical and computational information. A label table alone would hide this distinction; a YAML-backed release keeps it available for inspection, threshold analysis and learning tasks. The example metadata are listed in the Supplementary Information, Sec.~S4.

The spectrum-resolved design also defines how the data can be reused. A classifier may learn the PhononBench-defined Stable/unstable target. A regression model may predict $\wmin$ to distinguish near-threshold modes from deep imaginary branches. A calibration study may move the $-10^{-3}$ THz threshold and quantify label changes. These tasks all refer back to the same released YAML objects rather than to disconnected post-processing tables.

\section{Chemical and structural coverage}

\begin{figure}[H]
  \centering
  \includegraphics[width=0.98\textwidth]{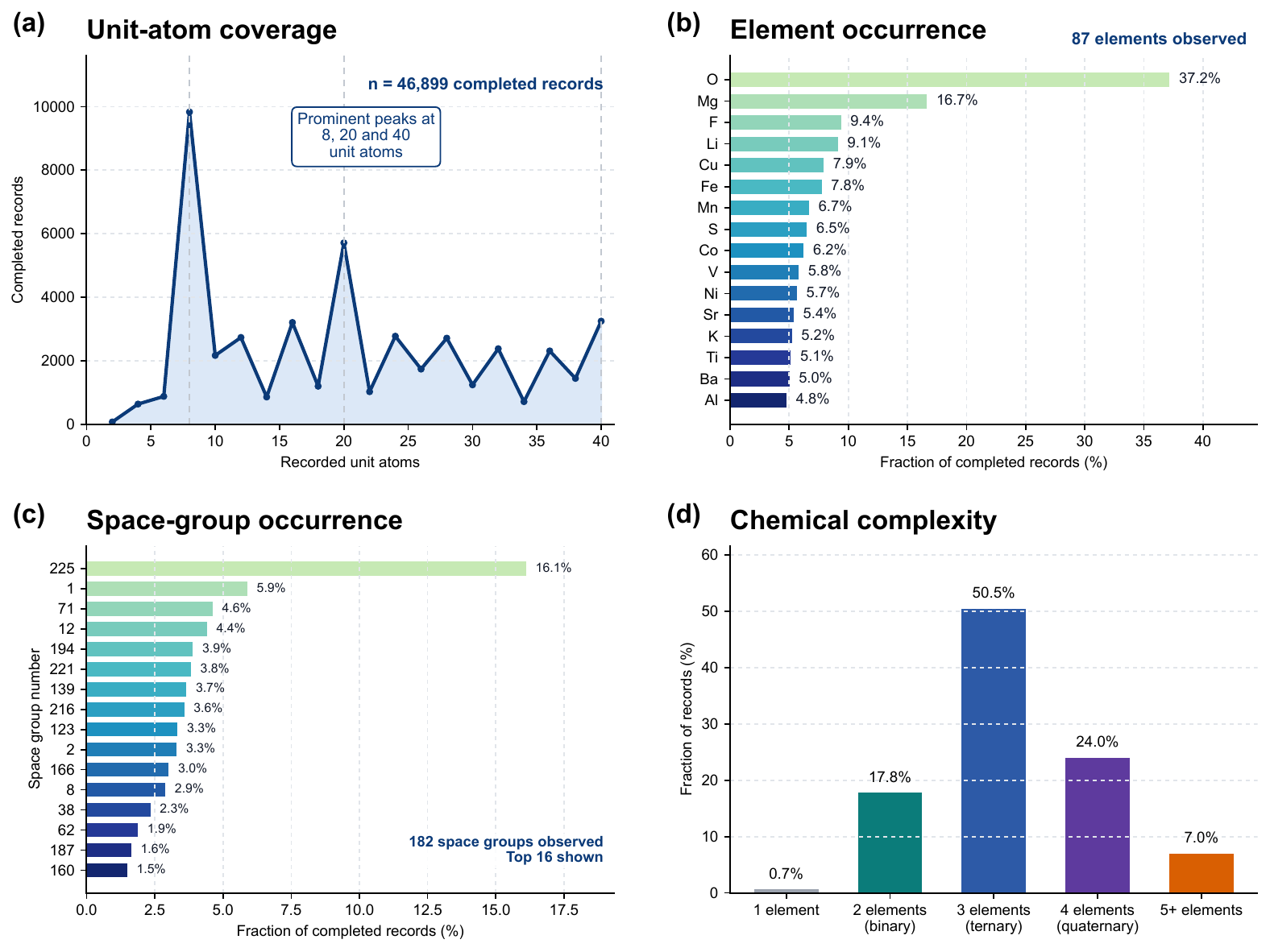}
  \caption{\label{fig:coverage}\textbf{Chemical and structural coverage of the completed cohort.} All panels are computed over the 46,899 completed label+YAML records. \textbf{a,} Distribution of recorded unit-cell atom counts within the MP40 scope, with prominent peaks at 8, 20 and 40 atoms. \textbf{b,} Element occurrence, counted once per completed record when an element appears in the parsed formula. \textbf{c,} Ranked space-group occurrence, showing broad but uneven crystallographic coverage. \textbf{d,} Chemical complexity measured by the number of distinct elements in the parsed formula, with ternary and quaternary compounds forming the largest groups.}
\end{figure}

Figure~\ref{fig:coverage} summarizes the completed cohort rather than the planned task list. All panels use the same denominator: the 46,899 completed label+YAML records. The completed cohort spans 87 observed elements and 182 observed space groups. Oxygen appears in 37.2\% of completed records, making it the dominant element in the occurrence count. This O-rich character is chemically natural for MP-derived inorganic crystals: O$^{2-}$ forms robust coordination polyhedra with many metal cations, stabilizes a wide range of oxide and oxyanion frameworks, and supports charge-balanced multication chemistries. The extended element, space-group and chemical-complexity statistics are reported in the Supplementary Information, Sec.~S6.

The chemical-complexity distribution is dominated by ternary and quaternary compounds: 50.5\% of records contain three elements and 24.0\% contain four. This distribution is relevant for reuse and model evaluation. Binary compounds provide important prototypes, but additional cation or anion species expand charge-compensation routes, site-substitution degrees of freedom and coordination environments. In oxides and oxyanion frameworks, for example, robust oxygen coordination can coexist with multiple metal sublattices, substitutions or charge-balancing species. For machine learning, the dataset therefore extends beyond simple binary compounds and contains many chemically richer compositions for which formula-level cues alone are less likely to capture the relevant lattice-dynamical behavior.

The unit-atom and space-group panels should be read as coverage statistics of the completed MP40 cohort, not as stability trends. Peaks at 8, 20 and 40 atoms reflect the MP40 source space, conventional-cell choices and crystallographic representation. The 40-atom edge is especially important: it admits ordered superstructures and multicomponent inorganic cells that are more demanding than very small-cell benchmarks, while avoiding a release dominated by very large phonon supercells. The ranked space-group plot shows broad but uneven crystallographic coverage, with high-symmetry prototypes and lower-symmetry distorted or ordered structures both present. Reporting these distributions is part of the audit: users can see the chemical and structural support over which the stability labels and spectra are defined.

\section{Technical validation}

\dataset\ is validated as an auditable workflow dataset rather than as a claim of model-independent absolute dynamical stability. The first validation layer is cohort accounting: task records, recovered local labels and recovered YAML outputs are intersected explicitly, giving 46,899 completed label+YAML records and identifying 1,067 relaxation failures outside the completed phonon denominator. This accounting prevents failed relaxations from being silently merged with completed imaginary-mode spectra.

The second validation layer is label traceability. For each completed record, the released label is derived from the sampled high-symmetry-path frequencies in the corresponding phonopy YAML file. The label rule is deterministic: any sampled band frequency below $-10^{-3}$ THz gives the completed-phonon \Unstable\ label, while all other completed spectra are labeled \Stable. Because the YAML file is released, users can re-extract $\wmin$, reproduce the binary label and test alternative thresholds without relying on a disconnected label table.

The third validation layer is distributional and physical sanity checking. The release reports the completed-cohort element distribution, space-group distribution, unit-atom distribution and chemical-complexity distribution using the same 46,899-record denominator. Representative spectra in Fig.~\ref{fig:spectra} further check that the label rule distinguishes stable spectra, shallow soft-mode spectra and deep-imaginary spectra in a physically interpretable way. Additional validation and recommended integrity checks are described in the Supplementary Information, Secs.~S6 and S10.

\section{Benchmark tasks and reuse}

\dataset\ is useful as a data release because spectra, labels and accounting are distributed together. It can also be used as a benchmark target because the completed cohort and label rule are fixed. For users who wish to compare models under a shared setting, we provide a deterministic formula-group reference split rather than the only valid way to use the release. The split assigns 37,633 training records, 4,724 validation records and 4,542 test records so that every one of the 35,348 formula groups appears in exactly one partition. This prevents exact-formula leakage and gives future studies a common starting point. If external materials databases are used for pretraining or feature generation, overlap with test structures, formulas or Materials Project identifiers should be reported because it changes how model scores should be interpreted. Split construction and reporting suggestions are detailed in the Supplementary Information, Secs.~S5 and S7.

The most direct benchmark task is PhononBench-defined stability classification on the completed label+YAML cohort. Since the completed cohort is imbalanced, with 30,216 completed-phonon \Unstable\ records and 16,683 \Stable\ records, accuracy alone is not a sufficient metric. Balanced accuracy, macro-F1, AUROC, AUPRC and class-specific recall are more informative for future comparisons. The released spectra also support $\wmin$ prediction and threshold-dependent relabeling. Separately, the 1,067 relaxation failures define a failure-aware triage problem: predicting whether a workflow produces an auditable spectrum is useful, but it should not be confused with predicting a completed phonon instability.

For reproducible baseline reporting, we recommend that future model reports include at least one simple composition-only baseline, one structure-aware baseline and, when computationally feasible, a graph-neural-network baseline such as CGCNN, MEGNet or ALIGNN \cite{Xie2018CGCNN,Chen2019MEGNet,Choudhary2021ALIGNN}. The purpose of these baselines is not to rank all model families in the present data paper, but to make the benchmark entry point clear: the split, labels, frequency target and failure-aware task are fixed, while methods can be compared under transparent reporting rules. The Supplementary Information, Sec.~S7, lists the recommended baseline protocol and metrics.

This benchmark use should be interpreted with the right boundary. \dataset\ does not claim model-independent absolute dynamical stability. It fixes a documented PhononBench/MatterSim--phonopy reference workflow. Other potentials, denser sampling, larger supercells, anharmonic treatments or first-principles phonon calculations may change borderline spectra. That limitation is also why the YAML object is released: users can inspect imaginary branches, measure threshold sensitivity and validate high-confidence candidates with the workflow or higher-fidelity calculations when needed. Additional interpretation boundaries are summarized in the Supplementary Information, Sec.~S8.

\section{Relation to machine-learning targets}

Materials machine learning increasingly relies on fixed tasks, transparent splits and reproducible metrics, as illustrated by Matbench, Matbench Discovery and OC20 \cite{Dunn2020Matbench,Riebesell2025MatbenchDiscovery,Chanussot2021OC20}. \dataset\ brings the same logic to workflow-defined phonon stability while retaining the spectral evidence behind each target. Descriptor-based materials learning, matminer workflows, crystal graph neural networks, SchNet, MEGNet and ALIGNN are natural model families for future structure-to-label and structure-to-frequency studies \cite{Ward2016GeneralML,Ward2018Matminer,Xie2018CGCNN,Schutt2018SchNet,Chen2019MEGNet,Choudhary2021ALIGNN}. Universal interatomic potentials such as M3GNet, CHGNet, NequIP, MACE and MatterSim have made large atomistic workflows increasingly practical \cite{Chen2022M3GNet,Deng2023CHGNet,Batzner2022NequIP,Batatia2022MACE,Batatia2025MACEFoundation,Yang2024MatterSim}. Recent work also shows that phonons remain a demanding and informative test for such models \cite{Loew2025MLIPPhonons}. \dataset\ gives this ecosystem a concrete reference target for phonon-stability screening: not an informal pass/fail label, but a spectrum-backed workflow result. The present release does not rank model families; it supplies the shared target, split files and spectral evidence needed for future comparisons.

\section{Usage notes and limitations}

The release is intended for three common entry points. First, users can read the completed table, follow the \texttt{yaml\_relative\_path} field to a phonopy YAML file, reconstruct the released dispersion and reproduce the label rule. Second, benchmark users can join the completed table with the reference split table and evaluate models on the fixed formula-group split. Third, workflow users can combine the completed table and the failure table to study whether a structure is likely to produce an auditable phonon spectrum under the reference route.

Several limitations should be retained in downstream use. \dataset\ is not a universal statement about thermodynamic stability, finite-temperature stability or DFT-level phonon stability. It records the result of a specified MP-derived PhononBench/MatterSim--phonopy workflow, a specified supercell choice, a specified path sampling procedure and a specified numerical threshold. Borderline structures, especially those with shallow soft modes, should be treated as candidates for follow-up calculation rather than final physical classifications. The released spectra and failure accounting are meant to make these limitations measurable instead of hidden.

\section{Conclusion}

\dataset\ converts a large MP-derived PhononBench calculation into a reusable spectrum-resolved benchmark dataset for workflow-defined phonon stability. The main release contains 46,899 completed label+YAML records, including 16,683 \Stable\ records and 30,216 completed-phonon \Unstable\ records, with 1,067 relaxation failures reported separately. Its central contribution is the release object: labels, spectra, splits and failure accounting are kept together, so each completed stability decision can be traced back to phonon evidence. This makes \dataset\ useful for stability classification, minimum-frequency analysis, threshold-dependent relabeling and failure-aware screening, while making the reference workflow and its boundaries explicit.

\section*{Data availability statement}

The dataset is being archived in Science Data Bank at \url{https://www.scidb.cn/detail?dataSetId=10.57760/sciencedb.38735}; the DOI \url{https://doi.org/10.57760/sciencedb.38735} should be cited once registration is active. The Science Data Bank record is intended to be the archival source for the dataset. The companion GitHub repository at \url{https://github.com/DreamLufei/phononbench_mp40} provides the calculation workflow, scripts and lightweight access utilities, but does not replace the archived dataset record. Users should report the data record used for the dataset and the Git commit used for code-level reuse. Release-file definitions, recommended integrity checks and example access steps are provided in the Supplementary Information.

\section*{Acknowledgements}
The work is supported by the National Natural Science Foundation of China (Nos. 62476278 and 11934020), Beijing Natural Science Foundation (No. Z250005) and the National Key R\&D Program of China (Grant No. 2024YFA1408601). Computational resources were provided by the Physical Laboratory of High Performance Computing at Renmin University of China.

\clearpage
\bibliographystyle{unsrt}
\bibliography{references}

\clearpage
\begin{center}
{\Large\bfseries Supplementary Information}\\[0.6em]
{\large \dataset: a spectrum-resolved benchmark dataset for phonon stability}\\[0.8em]
Wen-Kao Li$^{1,\dagger}$, Ze-Feng Gao$^{1,\dagger,*}$, Zhong-Yi Lu$^{1,*}$\\[2pt]
\small $^1$School of Physics and Key Laboratory of Quantum State Construction and Manipulation (Ministry of Education), Renmin University of China, Beijing 100872, China\\[2pt]
\small ${}^{\dagger}$These authors contributed equally.\\
\small ${}^{*}$e-mail: zfgao@ruc.edu.cn; zlu@ruc.edu.cn
\end{center}
\vspace{1em}
\section*{S1. Data schema and release files}

High-level cohort accounting is summarized in Table~1 of the main text. This section defines the release files used to produce those counts and the manuscript figures. The release is organized around four tabular data objects, release metadata and the corresponding phonopy YAML outputs. The master table records the audited MP40 task scope. The completed table contains only entries with both a local workflow label and a matching local YAML output. The failure table records relaxation failures without matching YAML outputs. The split table gives the reference train, validation and test assignment for completed records. The metadata and checksum files, when provided, define the release version and support file-integrity checks after download.

The MP40 scope denotes recorded cells with $N_{\mathrm{atoms}}\le 40$. This range was used to balance chemical and structural coverage against the cost of systematic phonon calculations. Under the $2\times2\times2$ supercell used in the workflow, a 40-atom recorded cell corresponds to an up-to-320-atom force-evaluation cell before symmetry reduction.

\begin{center}
\small
\textbf{Table S1. Release files used for the manuscript and release definition.}
\vspace{0.4em}

\begin{tabular}{lll}
\toprule
Data object & Records & Role \\
\midrule
Master table & 47,969 & \parbox[t]{0.42\textwidth}{Audited task-level MP40 cohort} \\
Completed table & 46,899 & \parbox[t]{0.42\textwidth}{Completed label+YAML release cohort} \\
Failure table & 1,067 & \parbox[t]{0.42\textwidth}{Relaxation failures without local YAML} \\
Split table & 46,899 & \parbox[t]{0.42\textwidth}{Reference split assignments for completed records} \\
Phonopy YAML spectra & 46,899 & \parbox[t]{0.42\textwidth}{Primary phonopy YAML spectra for completed records} \\
Metadata file & -- & \parbox[t]{0.42\textwidth}{Release version, creation date, repository information and license metadata, when provided} \\
Checksum file & -- & \parbox[t]{0.42\textwidth}{File-integrity checksums for release tables and metadata files, when provided} \\
\bottomrule
\end{tabular}
\end{center}

\section*{S2. Field definitions}

The common fields are listed in Table~S2.

\begin{center}
\small
\textbf{Table S2. Common field definitions.}
\vspace{0.4em}

\begin{tabular}{ll}
\toprule
Field & Definition \\
\midrule
\texttt{structure\_name} & \parbox[t]{0.62\textwidth}{Workflow key used to match task records, labels and YAML files.} \\
\texttt{source\_id} & \parbox[t]{0.62\textwidth}{First token of \texttt{structure\_name}, retained as a source identifier.} \\
\texttt{formula} & \parbox[t]{0.62\textwidth}{Formula token parsed from \texttt{structure\_name}.} \\
\texttt{space\_group} & \parbox[t]{0.62\textwidth}{Space-group number parsed from the \texttt{sg-} suffix.} \\
\texttt{unit\_atoms} & \parbox[t]{0.62\textwidth}{Recorded unit-cell atom count in the task table.} \\
\texttt{workflow\_label} & \parbox[t]{0.62\textwidth}{Raw local PhononBench label string, either \texttt{Stable} or \texttt{unStable}; the latter spelling is retained only for the raw workflow label.} \\
\texttt{completed\_label} & \parbox[t]{0.62\textwidth}{Manuscript label after enforcing the label+YAML completion rule.} \\
\texttt{omega\_min\_thz} & \parbox[t]{0.62\textwidth}{Lowest sampled high-symmetry-path phonon frequency, in THz, extracted from the corresponding YAML spectrum.} \\
\texttt{label\_threshold\_thz} & \parbox[t]{0.62\textwidth}{Numerical threshold used to derive the completed label from the sampled band frequencies.} \\
\texttt{yaml\_relative\_path} & \parbox[t]{0.62\textwidth}{Path to the local phonopy YAML output for completed records.} \\
\texttt{failure\_type} & \parbox[t]{0.62\textwidth}{Failure-accounting class for records outside the completed spectral cohort.} \\
\texttt{split}, \texttt{group\_key} & \parbox[t]{0.62\textwidth}{Reference split assignment and formula-level split group.} \\
\bottomrule
\end{tabular}
\end{center}

\section*{S3. Workflow and label rule}

The labels follow the local PhononBench workflow used for this release. Structures are relaxed before finite-displacement phonon calculations. The phonon calculation uses a $2\times2\times2$ supercell matrix, a displacement distance of 0.01~\AA, drift correction on displaced-supercell forces and symmetrized force constants. Band paths are generated through the seekpath/spglib convention with 101 points per segment.

For reproducibility, downstream users should report the data-release version, the Git commit used for parsing utilities, and the versions of the main phonon-analysis packages used in any reprocessing. The manuscript labels in the release are tied to the released YAML spectra and the threshold below; if a user regenerates phonons with a different potential, supercell, band path or threshold, the resulting labels should be reported as a new derived target rather than as the original \dataset\ label.

The completed stability label is derived from the high-symmetry-path frequencies. A structure is labeled \Unstable\ when any band frequency satisfies
\[
\omega < -10^{-3}\ {\rm THz}.
\]
Otherwise it is labeled \Stable. The release preserves the YAML spectrum so that users can reconstruct the band structure, extract $\wmin$, and apply stricter or looser thresholds.

\section*{S4. Representative spectra}

Three representative examples were selected to show distinct spectral regimes behind the binary labels.

\begin{center}
\small
\textbf{Table S3. Representative spectra shown in Fig.~2 of the main text.}
\vspace{0.4em}

\begin{tabular}{llll}
\toprule
Example & Space group & Label class & $\wmin$ (THz) \\
\midrule
MgB$_2$ & 191 & \Stable & 0.0000 \\
RbPbCl$_3$ & 221 & Soft-mode completed-phonon \Unstable & -1.4984 \\
BaTiO$_3$ & 221 & Deep-imaginary completed-phonon \Unstable & -7.9102 \\
\bottomrule
\end{tabular}
\end{center}

\section*{S5. Reference split}

The reference split is deterministic at the formula-group level. The group key is the parsed formula token. A SHA-256 hash of this key assigns each formula group to train, validation or test using the rule train $<0.80$, validation $<0.90$ and test otherwise. This split prevents exact-formula leakage across train, validation and test sets.

\begin{center}
\small
\textbf{Table S4. Reference split counts by completed label.}
\vspace{0.4em}

\begin{tabular}{lrrr}
\toprule
Split & \Stable & Completed-phonon \Unstable & Total \\
\midrule
Train & 13,304 & 24,329 & 37,633 \\
Validation & 1,711 & 3,013 & 4,724 \\
Test & 1,668 & 2,874 & 4,542 \\
\midrule
Total & 16,683 & 30,216 & 46,899 \\
\bottomrule
\end{tabular}
\end{center}

The completed cohort contains 35,348 formula groups. Under the reference split, every formula group appears in exactly one split, giving zero exact-formula leakage.

\section*{S6. Coverage statistics}

The completed cohort contains 87 observed elements and 182 observed space groups. The most frequent elements are O, Mg, F, Li, Cu, Fe, Mn, S, Co, V, Ni, Sr, K, Ti, Ba and Al. The corresponding occurrence percentages over completed records are 37.2\%, 16.7\%, 9.4\%, 9.1\%, 7.9\%, 7.8\%, 6.7\%, 6.5\%, 6.2\%, 5.8\%, 5.7\%, 5.4\%, 5.2\%, 5.1\%, 5.0\% and 4.8\%.

The leading space groups are 225, 1, 71, 12, 194, 221, 139, 216, 123, 2, 166, 8, 38, 62, 187 and 160. Their occurrence percentages are 16.1\%, 5.9\%, 4.6\%, 4.4\%, 3.9\%, 3.8\%, 3.7\%, 3.6\%, 3.3\%, 3.3\%, 3.0\%, 2.9\%, 2.3\%, 1.9\%, 1.6\% and 1.5\%.

Chemical complexity is measured by the number of distinct elements in the parsed formula. The completed cohort contains 0.7\% unary, 17.8\% binary, 50.5\% ternary, 24.0\% quaternary and 7.0\% five-or-more-element records.

\section*{S7. Benchmark uses, tasks and metrics}

The release supports several benchmark-style uses without prescribing a single field-wide protocol. For users who wish to compare models under a shared setting, the deterministic formula-group split provides a reference split rather than the only valid way to use the release. When another split is used, exact-formula leakage and overlap with external pretraining data should be reported.

\textit{Stability classification.} The input is a crystal structure in the completed label+YAML cohort. The target is the workflow-defined \Stable/\Unstable\ label. Recommended metrics include balanced accuracy, macro-F1, AUROC, AUPRC and class-specific recall. Accuracy alone is not sufficient because the completed cohort is label-imbalanced. Scores should be reported for the whole test set and for chemically meaningful subsets, including atom-count bins, chemical-complexity classes, oxygen-containing versus non-oxygen-containing records and common space-group families.

\textit{Minimum-frequency regression.} The input is the same completed cohort. The target is $\wmin$ extracted from the YAML spectrum. Recommended metrics include MAE and RMSE in THz, Spearman rank correlation and error stratified by near-threshold, soft-mode and deep-instability cases.

\textit{Failure-aware workflow triage.} The input includes completed records and the 1,067 relaxation failures. The target is whether a record produced an auditable local YAML spectrum. Recommended metrics include precision, recall and F1 for the failure class. This task should be reported separately from physical phonon-stability prediction.

\textit{Candidate validation.} Model predictions on new structures should be treated as screening outputs. A predicted stable candidate should be validated by rerunning the PhononBench workflow, and by a higher-fidelity phonon calculation when the downstream use requires stricter confirmation. This separates model scoring on the release from final materials validation.

\textit{Reporting checklist.} A transparent model report should state the structure representation, model family, training data, external pretraining data, split used for evaluation, subgroup scores when relevant, classification threshold, $\wmin$ regression metrics when used and the validation route for any newly proposed stable candidates.

\textit{Recommended baseline protocol.} A minimal benchmark report should include a majority-class baseline, a composition-only baseline and one structure-aware baseline. The composition-only baseline can use stoichiometric and elemental statistics features, while the structure-aware baseline can use graph, local-environment or symmetry-aware descriptors. For graph neural networks, CGCNN, MEGNet and ALIGNN are natural starting points. Baseline scores should be reported on the fixed test split and should not be tuned on the test set. If a baseline uses pretrained representations or external materials data, exact overlap with \dataset\ test formulas, source identifiers or structures should be reported.

\section*{S8. Boundaries of interpretation}

\dataset\ is a release of a defined computational workflow. The labels inherit the workflow settings, including relaxation protocol, interatomic potential, finite-displacement setup, band path and numerical threshold. Borderline structures may change under denser sampling, larger supercells, alternative potentials, anharmonic treatment or higher-fidelity follow-up calculations. The released spectra are intended to make these boundaries explicit: users can inspect the imaginary branch, measure the margin to the threshold and choose application-specific labels when needed.

\section*{S9. Usage notes}

The recommended first-use workflow is:
\begin{enumerate}
\item Download the archived release from Science Data Bank and verify the release tables against the accompanying checksum file, when provided.
\item Read the completed table and select one completed record.
\item Use \texttt{yaml\_relative\_path} to locate the corresponding phonopy YAML file.
\item Parse the YAML frequencies, compute the minimum sampled frequency $\wmin$ and apply the threshold $\omega < -10^{-3}$~THz to reproduce the completed label.
\item Join the completed table with the split table before model training so that train, validation and test records follow the reference formula-group split.
\end{enumerate}

For label-threshold studies, users should keep the original completed label unchanged and store any threshold-dependent relabeling in a new derived column. This makes it possible to compare alternative label conventions without losing the original workflow-defined target.

For failure-aware studies, users should combine the completed table with the failure table and define a separate binary target for whether a task produced an auditable YAML spectrum. This target should not be mixed with the completed-phonon Stable/unstable label.

\section*{S10. Technical validation checklist}

The following checks define the validation logic used for the release and are recommended for downstream mirrors of the data.

\begin{center}
\small
\textbf{Table S5. Validation and integrity checks.}
\vspace{0.4em}

\begin{tabular}{ll}
\toprule
Check & Purpose \\
\midrule
Record-count audit & \parbox[t]{0.62\textwidth}{Confirm the master, completed, failure and split counts used in the manuscript.} \\
Label+YAML intersection & \parbox[t]{0.62\textwidth}{Ensure that every completed record has both a recovered workflow label and a matching YAML spectrum.} \\
YAML path existence & \parbox[t]{0.62\textwidth}{Verify that every \texttt{yaml\_relative\_path} in the completed table resolves to a released YAML file.} \\
Threshold relabeling & \parbox[t]{0.62\textwidth}{Recompute $\wmin$ from the YAML frequencies and check that the threshold rule reproduces the completed label.} \\
Split leakage check & \parbox[t]{0.62\textwidth}{Confirm that every formula-group key appears in exactly one of train, validation and test.} \\
Distributional audit & \parbox[t]{0.62\textwidth}{Recompute unit-atom, element, space-group and chemical-complexity distributions over the completed denominator.} \\
Checksum verification & \parbox[t]{0.62\textwidth}{Confirm that downloaded tables and metadata match the archived release files.} \\
\bottomrule
\end{tabular}
\end{center}

\end{document}